%% file: main.tex
\documentclass[10pt,twocolumn,letterpaper]{article}

\pdfoutput=1

\usepackage[final,pagenumbers]{cvpr}

\input{preamble}

\definecolor{cvprblue}{rgb}{0.21,0.49,0.74}
\usepackage[breaklinks,colorlinks,allcolors=cvprblue]{hyperref}

\def\paperID{*****}
\def\confName{CVPR}
\def\confYear{2026}

\title{Zero-Shot Vision-Language Models for Classroom Engagement Recognition:\\
A Benchmark Study of Prompt Sensitivity and Cross-Dataset Generalization}

\author{%
Aman Goyal$^{*}$\\
Carnegie Mellon University
\and
Kshama Nitin Shah$^{*}$\\
University of Michigan, Ann Arbor
\and
Kemmannu Vineet Venkatesh Rao\\
Magna International
}

\begin{document}
\maketitle
\renewcommand{\thefootnote}{\fnsymbol{footnote}}
\footnotetext[1]{Equal contribution.}
\footnotetext[2]{Presented as a non-archival paper at the CV4Edu Workshop, CVPR 2026.}
\input{sec/abstract}
\input{sec/intro}
\input{sec/related}
\input{sec/method}
\input{sec/results}
\input{sec/discussion}
\input{sec/conclusion}

{
    \small
    \bibliographystyle{ieeenat_fullname}
    \bibliography{refs}
}

\input{sec/suppl}

\end{document}

%% file: preamble.tex
\usepackage{amsmath}
\usepackage{amssymb}
\usepackage{booktabs}
\usepackage{multirow}
\usepackage{graphicx}
\usepackage{xcolor}
\usepackage{subcaption}
\usepackage{microtype}
\usepackage{array}
\usepackage{tabularx}
\usepackage{colortbl}
\usepackage{xspace}

\providecommand{\eg}{\textit{e.g.}\xspace}

\providecommand{\etal}{\textit{et al.}\xspace}

\newcolumntype{C}[1]{>{\centering\arraybackslash}p{#1}}

\definecolor{rowgray}{gray}{0.95}
\definecolor{rowblue}{rgb}{0.90,0.93,0.98}

%% file: sec/abstract.tex
\begin{abstract}
Automated classroom engagement recognition holds substantial promise for
scalable learning analytics, yet the suitability of modern
Vision-Language Models (VLMs) for this task under zero-shot conditions
remains largely unexplored.  We present a systematic benchmark that
evaluates five widely-used VLMs---CLIP, BLIP-VQA, GPT-4o,
LLaVA-1.5-7B, Qwen2.5VL-7B-Instruct ---across two complementary educational datasets: DAiSEE,
an individual-student video dataset (300 sampled test clips), and the
Student Classroom Behaviour dataset (SCB, 1{,}168 scene-level images).
Each model is probed with three prompt variants spanning minimal,
rubric-anchored, and chain-of-thought designs.  Our experiments reveal
three primary failure modes of zero-shot VLMs for engagement recognition:
(1)~\emph{near-random performance on individual students}, with Cohen's
$\kappa$ never exceeding 0.10 on DAiSEE; (2)~\emph{severe class
collapse}, where models assign 85--100\% of predictions to a single
engagement level regardless of visual content; and (3)~\emph{extreme
prompt sensitivity}, with accuracy swings of up to 32 percentage points
on identical images depending solely on prompt phrasing.  Remarkably,
scene-level classification on SCB is substantially more tractable:
CLIP and GPT-4o achieve $\kappa \approx 0.60$ when prompted with
behaviorally-grounded rubrics.  We also document a practical barrier for
deployment---GPT-4o's safety filters reject 98\% of chain-of-thought
requests involving individual student faces.  Our findings provide a
calibrated baseline and surface critical design considerations for the
use of VLMs in educational observation systems.
\end{abstract}

%% file: sec/intro.tex
\section{Introduction}
\label{sec:intro}

Understanding student engagement is central to effective teaching.
Disengagement---whether manifested as wandering attention, passive
participation, or behavioural withdrawal---predicts lower learning
outcomes and higher attrition~\cite{fredricks2004school}.
Traditionally, measuring engagement at scale requires trained human
observers~\cite{pianta2008classroom}, a resource-intensive process that
is difficult to replicate in large lecture halls or remote learning
environments.  The rapid proliferation of cameras in educational
settings, coupled with powerful visual recognition models, has therefore
sparked interest in automated engagement sensing~\cite{whitehill2014faces,
gupta2016daisee,thomas2018predicting,liao2021deep}.

Contemporary Vision-Language Models (VLMs) represent a potential leap
forward for this problem.  Unlike supervised classifiers that require
large labelled corpora specific to each school or cultural context,
models such as CLIP~\cite{radford2021learning},
BLIP~\cite{li2022blip}, GPT-4o~\cite{openai2024gpt4o}, 
LLaVA~\cite{liu2023visual} and Qwen~\cite{bai2025qwen25vl}can be queried zero-shot via natural language,
with no task-specific fine-tuning.  Their ability to reason from open
prompts makes them attractive as \emph{plug-and-play} educational
observers.  Yet, to our knowledge, no prior work has systematically
benchmarked VLMs for classroom engagement recognition under controlled,
zero-shot conditions.

\begin{figure}[t]
    \centering
    \includegraphics[width=\linewidth]{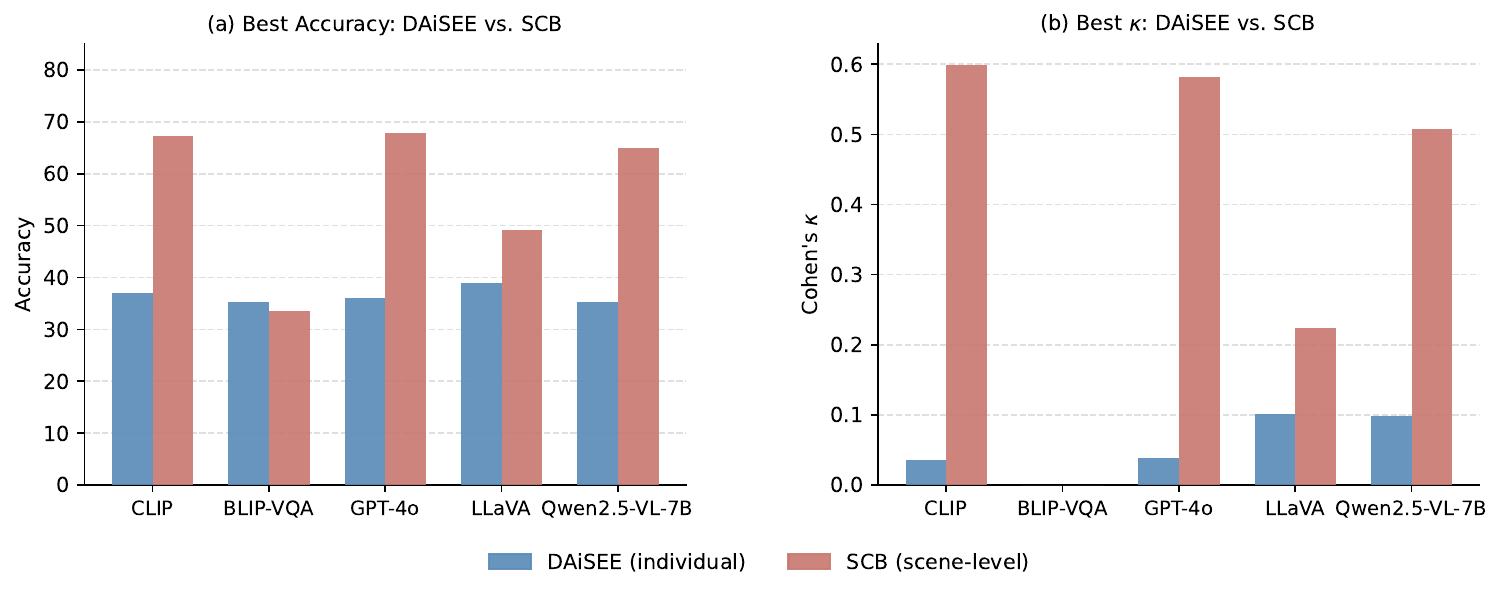}
    \caption{Best zero-shot accuracy and Cohen's $\kappa$ per model on
    DAiSEE (individual student, video frames) versus SCB (classroom
    scene images).  Scene-level classification is substantially more
    tractable, with CLIP and GPT-4o achieving $\kappa\!\approx\!0.60$
    on SCB while remaining near-random on DAiSEE.}
    \label{fig:cross_dataset}
\end{figure}

\paragraph{Our contributions.}  This paper presents the following:

\begin{itemize}
  \item A zero-shot benchmark of five VLMs on two complementary
    datasets: the individual-student DAiSEE dataset~\cite{gupta2016daisee}
    and the scene-level Student Classroom Behaviour (SCB)
    dataset~\cite{wang2023scb}, using three systematically-designed
    prompt variants (minimal, rubric, chain-of-thought).

  \item A rigorous quantitative analysis documenting class collapse,
    near-random agreement on individual-level data ($\kappa \leq 0.10$),
    and substantial prompt sensitivity (accuracy variance up to 32 pp).

  \item The finding that scene-level engagement classification is
    significantly more tractable for VLMs, with CLIP achieving
    $\kappa = 0.60$ and GPT-4o achieving $\kappa = 0.58$ on SCB.

  \item Documentation of GPT-4o's safety-guardrail refusals blocking
    chain-of-thought engagement analysis for individual student faces,
    with practical implications for educational AI deployment.

  \item A self-consistency analysis showing that generative VLMs
    (GPT-4o, LLaVA) change their predictions for 27--32\% of images
    across repeated runs at temperature $T\!=\!0.7$, raising reliability
    concerns for classroom monitoring applications.
\end{itemize}

Our results paint a nuanced picture: while VLMs are not yet reliable
individual engagement observers, they demonstrate meaningful capability
at the scene level.  This gap between individual and scene-level
performance has direct implications for system design---aggregating
information spatially before querying a VLM may be a more productive
paradigm than per-student analysis.

%% file: sec/related.tex
\section{Related Work}
\label{sec:related}

\paragraph{Engagement detection in educational settings.}
Early work by Whitehill \etal~\cite{whitehill2014faces} demonstrated
that facial features encode cognitive states relevant to learning.
The release of DAiSEE~\cite{gupta2016daisee} established the first large
publicly-available benchmark for four-level engagement classification
from student video.  Subsequent supervised methods leveraged spatiotemporal
features from video~\cite{liao2021deep}, recurrent networks over facial
action units~\cite{thomas2018predicting}, and attention-based
architectures~\cite{huang2019fine}.  More recently, spatiotemporal and attention-based
models trained on DAiSEE have reported accuracy of 55--65\%~\cite{liao2021deep,huang2019fine}, underscoring
the difficulty even for task-specific supervised models.
Classroom-level engagement has received comparatively less attention,
though several works address student behaviour recognition from overhead
cameras~\cite{wang2023scb,kaur2018classroom}.

\paragraph{Zero-shot visual recognition with VLMs.}
CLIP~\cite{radford2021learning} pioneered zero-shot image classification
via image-text cosine similarity, demonstrating competitive performance
on many standard benchmarks.  BLIP~\cite{li2022blip} introduced
bootstrapped image-text alignment that enabled visual question answering
without task-specific training.  Instruction-tuned generative models
such as LLaVA~\cite{liu2023visual} and GPT-4V/4o~\cite{openai2024gpt4o}
further extend this to open-ended visual reasoning.  A growing body of
work has applied these models to medical imaging~\cite{zhang2023pmc},
document understanding, and remote sensing, but their application to
affect recognition and educational observation has not been
systematically studied.

\paragraph{Prompt design and sensitivity.}
The sensitivity of VLMs to prompt phrasing has been studied in the
context of natural language processing~\cite{zhao2021calibrate,
lu2022fantastically}, but visual prompting for classification remains
less understood.  Studies on zero-shot classification~\cite{menon2022visual}
demonstrate that performance can vary by 10--30 percentage points across
prompt phrasings---a phenomenon we observe and quantify here.

%% file: sec/method.tex
\section{Methodology}
\label{sec:method}

\subsection{Datasets}

\paragraph{DAiSEE.}
The Dataset for Affective States in E-Environments
(DAiSEE)~\cite{gupta2016daisee} contains 9{,}068 ten-second video clips
from 112 subjects recorded via webcam during e-learning sessions.  Each
clip is annotated on a four-point Likert scale for engagement (0 = not
engaged, 3 = highly engaged) along with boredom, confusion, and
frustration.  We focus exclusively on engagement.  The official test
split comprises 1{,}784 clips across 21 subjects; the label distribution
is heavily skewed, with only 4 samples at level 0 and approximately 850
each at levels 2 and 3.

We evaluate on a stratified sample of $N\!=\!300$ test clips, retaining
all 4 level-0 clips, all 84 level-1 clips, and randomly sampling 106
clips each from levels 2 and 3.  From each 10-second clip, we extract
a single representative frame at $t\!=\!5$\,s (the temporal midpoint)
using \texttt{ffmpeg}.  This setup tests whether static visual cues
visible in a single webcam frame are sufficient for zero-shot engagement
inference.

\paragraph{Student Classroom Behaviour (SCB).}
The SCB dataset~\cite{wang2023scb} contains classroom scene images with
per-student bounding box annotations across six behaviour categories:
\emph{reading}, \emph{writing}, \emph{raising hand}, \emph{using phone},
\emph{bowing head}, and \emph{leaning over table}.  We use a combined
split of 1{,}168 validation images drawn from both the original 5K
corpus and an additional university subset.

Following the aggregation strategy proposed in the dataset documentation,
we derive a scene-level engagement label from the per-student annotations.
Each behaviour is mapped to an engagement score (0--3): disruptive
behaviours (\emph{phone}, \emph{bowing head}) score 0; passive
disengagement (\emph{leaning over}) scores 1; productive behaviours
(\emph{reading}, \emph{writing}) score 2; and active participation
(\emph{raising hand}) scores 3.  The scene-level label is the
mean across all annotated students, discretised via uniform thresholds
($\left[0,\,0.75\right) \to 0$; $\left[0.75,\,1.75\right) \to 1$;
$\left[1.75,\,2.75\right) \to 2$; $\left[2.75,\,3\right] \to 3$).
The resulting label distribution is highly imbalanced: 54.5\% of images
fall at level 3, 33.6\% at level 2, 10.1\% at level 1, and 1.9\%
at level 0.

\paragraph{Task formulation.}
We note that our evaluation of SCB differs fundamentally from prior work 
on this dataset. Existing methods~\cite{wang2023scb,teotia2024} 
treat SCB as a \emph{per-student object detection} task: they first localise 
individual students via bounding-box detectors (e.g., YOLOv5~\cite{wang2023scb}) 
and then classify each detected student's behaviour independently. In contrast, 
we formulate engagement recognition as \emph{scene-level image classification}, 
where the model receives the full classroom image and produces a single holistic 
engagement score aggregated across all visible students. This reformulation is 
deliberate: zero-shot VLMs are not designed for spatial object detection, and 
our results (Sec.~\ref{sec:scb_results}) demonstrate that scene-level 
cues---spatial layout, group dynamics, and overtly visible 
behaviours---are precisely the signals that make zero-shot engagement 
recognition tractable ($\kappa \approx 0.60$). Because these are fundamentally 
different task formulations operating at different output granularities, direct 
numerical comparison between our scene-level accuracy and prior per-student 
detection metrics is not meaningful. 

\subsection{Models}

We evaluate four models at substantially different points on the
capability-cost spectrum:

\textbf{CLIP (ViT-B/32)}~\cite{radford2021learning} is a discriminative
contrastive model.  Classification proceeds by computing cosine
similarity between the image embedding and embeddings of textual
descriptions of each engagement level; the class with the highest
similarity is the prediction.  Inference is fully deterministic.

\textbf{BLIP-VQA (blip-vqa-base)}~\cite{li2022blip} is a discriminative
VQA model.  We pose a natural-language question about student engagement;
the model generates a short free-text answer (\eg, ``not very'',
``yes'') that is mapped to a level via a hand-crafted lookup table.
Beam decoding with $k\!=\!3$ ensures deterministic output.

\textbf{GPT-4o}~\cite{openai2024gpt4o} is a proprietary closed-source
multimodal model accessed via the OpenAI API.  Images are submitted as
base64-encoded JPEG at ``low'' detail resolution.  Temperature is 0.0
for run 1 and 0.7 for subsequent consistency runs.

\textbf{LLaVA-1.5-7B}~\cite{liu2023visual} is an open-source instruction-tuned
model served locally through Ollama at 4-bit quantisation (reducing
memory from ${\sim}14$\,GB to ${\sim}4$\,GB) on a local machine with 16\,GB RAM.
We use the same prompts as GPT-4o to enable direct comparison.
Temperature follows the same protocol as GPT-4o.

\textbf{Qwen2.5-VL-7B-Instruct}~\cite{bai2025qwen25vl} is an open-source vision-language
model from Alibaba's Qwen team, loaded in \texttt{bfloat16} precision via
\texttt{Qwen2\_5\_VLForConditionalGeneration} from the Hugging Face Transformers
library~\cite{wolf2020transformers}. We use the same three prompt variants as GPT-4o
and LLaVA to enable direct comparison.  Temperature follows the same
protocol: $T\!=\!0.0$ for run~1 (deterministic) and $T\!=\!0.7$ for
self-consistency runs~2 and~3.

\subsection{Prompt Variants}

Three prompts are applied to each model to quantify sensitivity:

\textbf{P1 (Minimal)} provides only a four-label numeric scale with no
contextual guidance.  It establishes a lower bound on what each model
can infer from minimal language grounding.

\textbf{P2 (Rubric-Anchored)} augments P1 with explicit behavioural
anchors for each level, \eg, \emph{``looking away, yawning''} for
level 0 and \emph{``leaning forward, showing curiosity''} for level 3.
This mirrors how human annotators are typically trained.

\textbf{P3 (Chain-of-Thought)} instructs the model to first describe
the student's or class's observed facial expression and body posture
before committing to a rating.  This structured reasoning format has
shown benefits in language-only tasks~\cite{wei2022chain} and is
evaluated here in the visual domain.

For CLIP, the prompts are class-specific text templates.  For BLIP-VQA,
the prompts are question strings.  For GPT-4o and LLaVA, identical
prompt text is used to enable direct cross-model comparison.

\subsection{Evaluation Metrics}

All experiments are evaluated with four metrics: (i)~\textbf{Accuracy}
(exact-match 4-class); (ii)~\textbf{F1 Macro}, the unweighted mean of
per-class F1 scores, which penalises class collapse; (iii)~\textbf{Cohen's
$\kappa$ (quadratic weighted)}, which accounts for ordinal agreement
structure and penalises predictions far from ground truth more severely;
and (iv)~\textbf{MSE} (mean squared error), measuring average ordinal
distance.  For self-consistency, we report full 3-run agreement rate
and pairwise agreement rate.

%% file: sec/results.tex
\section{Experimental Results}
\label{sec:results}

\subsection{DAiSEE: Individual-Student Engagement}
\label{sec:daisee_results}

Table~\ref{tab:daisee_main} shows the best-performing configuration for
each model alongside published supervised baselines.  The gap is
striking: even the strongest VLM achieves only 39.0\% accuracy and
$\kappa\!=\!0.10$ (LLaVA with chain-of-thought), while supervised
convolutional+temporal models reach 58.8\%~\cite{liao2021deep} and
context-aware transformers achieve $\kappa > 0.35$~\cite{huang2019fine}.

\begin{table}[t]
\centering
\small
\caption{Best classification results on DAiSEE ($N\!=\!300$ test clips).
  Supervised baselines are drawn from published work on the full test set.
\\  Best VLM results are highlighted in \textbf{bold}.}
\label{tab:daisee_main}
\setlength{\tabcolsep}{4pt}
\begin{tabular}{@{}lcccc@{}}
\toprule
Model & Acc.\ (\%) & F1 Macro & $\kappa$ & MSE \\
\midrule
\rowcolor{rowgray}
\multicolumn{5}{l}{\textit{Supervised baselines (literature)}} \\
ResNet+TCN~\cite{liao2021deep}        & 58.8 & -- & --  & 0.18 \\
Huang \etal~\cite{huang2019fine}      & 62.3 & -- & 0.37 & -- \\
CavT~\cite{abedi2021affect}           & --   & -- & --  & 0.038 \\
\midrule
\rowcolor{rowblue}
\multicolumn{5}{l}{\textit{Zero-shot VLMs (this work)}} \\
CLIP (P1)                             & 37.0 & 0.224 & 0.036 & 1.00 \\
BLIP-VQA (P3)                         & 35.3 & 0.131 & 0.000 & 0.69 \\
GPT-4o (P3)$^\dagger$                 & 36.0 & 0.153 & 0.039 & 0.69 \\
\textbf{LLaVA-1.5 (P3)}               & \textbf{39.0} & \textbf{0.210} & \textbf{0.101} & \textbf{0.72} \\
Qwen2.5-VL-7B-Instruct (P3)
 & 35.3 & 0.136 & $-$0.041 & 0.737 \\

\bottomrule
\end{tabular}
\vspace{1pt}
\begin{minipage}{\linewidth}
  \scriptsize $^\dagger$98\% of GPT-4o P3 responses were safety refusals;
  results reflect fallback-to-level-2 parsing.
\end{minipage}
\end{table}

Table~\ref{tab:daisee_allprompts} presents the full prompt-by-prompt
breakdown.  Several patterns stand out.  First, no single model-prompt
combination surpasses $\kappa\!=\!0.11$, confirming near-random
ordinal agreement across the board.  Second, all models show accuracy
near or below the majority-class baseline of 35.3\% (level 2),
suggesting the higher accuracies reflect correct predictions of the
modal class rather than genuine four-class discrimination.

\begin{table}[t]
\centering
\small
\caption{Full prompt-sensitivity results on DAiSEE.  Pred.\ Dist.\ shows
  the fraction of predictions in each class (GT: 1\%--28\%--35\%--35\%).}
\label{tab:daisee_allprompts}
\setlength{\tabcolsep}{3pt}
\begin{tabular}{@{}llcccc@{}}
\toprule
Model & Prompt & Acc. & F1 & $\kappa$ & MSE \\
\midrule
\multirow{3}{*}{CLIP}
  & P1 & 37.0 & 0.224 & 0.036 & 1.00 \\
  & P2 & 31.3 & 0.184 & 0.068 & 1.39 \\
  & P3 & 35.0 & 0.231 & $-$0.042 & 1.21 \\
\midrule
\multirow{3}{*}{BLIP-VQA}
  & P1 & 25.3 & 0.116 & $-$0.018 & 2.39 \\
  & P2 & 32.7 & 0.153 & 0.058 & 0.94 \\
  & P3 & 35.3 & 0.131 & 0.000 & 0.69 \\
\midrule
\multirow{3}{*}{GPT-4o}
  & P1 & 27.7 & 0.144 & 0.067 & 1.67 \\
  & P2 & 32.3 & 0.192 & 0.075 & 1.51 \\
  & P3$^\dagger$ & 36.0 & 0.153 & 0.039 & 0.69 \\
\midrule
\multirow{3}{*}{LLaVA-1.5}
  & P1 & 5.3 & 0.047 & 0.037 & 4.07 \\
  & P2 & 15.7 & 0.097 & 0.045 & 2.84 \\
  & \textbf{P3} & \textbf{39.0} & \textbf{0.210} & \textbf{0.101} & \textbf{0.72} \\
\midrule
\multirow{3}{*}{Qwen2.5-VL-7B-Instruct }
  & P1 & 30.0 & 0.25 & 0.098 & 1.34 \\
  & P2 & 30.0 & 0.257 & 0.097 & 1.30 \\
  & P3 & 35.3 & 0.136 & $-$0.041 & 0.737 \\
\bottomrule
\end{tabular}
\vspace{1pt}
\begin{minipage}{\linewidth}
  \scriptsize $^\dagger$Safety refusal rate $\approx$98\%; level-2 fallback inflates accuracy.
\end{minipage}
\end{table}

\subsection{SCB: Scene-Level Classroom Engagement}
\label{sec:scb_results}

Table~\ref{tab:scb_main} shows results on the SCB dataset.  Performance
improves dramatically compared with DAiSEE.  CLIP with chain-of-thought
templates reaches $\kappa\!=\!0.60$ and GPT-4o with the rubric prompt
reaches $\kappa\!=\!0.58$---placing both in the ``moderate'' agreement
range~\cite{landis1977measurement}.  These numbers approach the
performance of supervised methods in comparable scene-understanding tasks.

\begin{table}[t]
\centering
\small
\caption{Results on SCB ($N\!=\!1{,}168$ validation images).
  All numbers are the best single prompt per model.}
\label{tab:scb_main}
\setlength{\tabcolsep}{4pt}
\begin{tabular}{@{}lcccc@{}}
\toprule
Model & Best Prompt & Acc.\ (\%) & $\kappa$ & MSE \\
\midrule
CLIP (ViT-B/32)  & P3 & 67.3 & 0.599 & 0.586 \\
BLIP-VQA         & P1/P2/P3$^*$ & 33.6 & 0.000 & 0.721 \\
\textbf{GPT-4o}  & \textbf{P2} & \textbf{67.9} & \textbf{0.582} & \textbf{0.372} \\
LLaVA-1.5        & P2 & 49.2 & 0.224 & 2.666 \\
Qwen2.5-VL-7B-Instruct        & P2 & 64.9 & 0.508 & 0.408 \\
\bottomrule
\end{tabular}
\vspace{1pt}
\begin{minipage}{\linewidth}
  \scriptsize $^*$BLIP-VQA produces identical predictions across all
  three prompts on SCB (complete zero sensitivity; see Sec.~\ref{sec:analysis}).
\end{minipage}
\end{table}

The full SCB prompt breakdown (Table~\ref{tab:scb_allprompts}) reveals
interesting cross-prompt dynamics.  CLIP shows a dramatic positive
response to the emotional/behavioural template in P3 (from $\kappa\!=\!0.10$
at P1 to $\kappa\!=\!0.60$ at P3).  GPT-4o peaks with the rubric prompt
(P2) but drops 32 pp in accuracy under chain-of-thought (P3).  LLaVA
performs best under P2 on SCB, unlike on DAiSEE where P3 is superior.

\begin{table}[t]
\centering
\small
\caption{Full prompt-sensitivity results on SCB.}
\label{tab:scb_allprompts}
\setlength{\tabcolsep}{3pt}
\begin{tabular}{@{}llcccc@{}}
\toprule
Model & Prompt & Acc. & F1 & $\kappa$ & MSE \\
\midrule
\multirow{3}{*}{CLIP}
  & P1 & 14.4 & 0.097 & 0.100 & 1.926 \\
  & P2 & 24.7 & 0.172 & 0.263 & 2.564 \\
  & P3 & 67.3 & 0.472 & 0.599 & 0.586 \\
\midrule
\multirow{3}{*}{BLIP-VQA}
  & P1 & 33.6 & 0.126 & 0.000 & 0.721 \\
  & P2 & 33.6 & 0.126 & 0.000 & 0.721 \\
  & P3 & 33.6 & 0.126 & 0.000 & 0.721 \\
\midrule
\multirow{3}{*}{GPT-4o}
  & P1 & 39.1 & 0.211 & 0.154 & 0.688 \\
  & P2 & 67.9 & 0.400 & 0.582 & 0.372 \\
  & P3 & 51.8 & 0.273 & 0.260 & 0.603 \\
\midrule
\multirow{3}{*}{LLaVA-1.5}
  & P1 & 7.4  & 0.067 & 0.027 & 5.343 \\
  & P2 & 49.2 & 0.353 & 0.224 & 2.666 \\
  & P3 & 33.9 & 0.131 & $-$0.027 & 0.762 \\
\midrule
  \multirow{3}{*}{Qwen2.5-VL-7B-Instruct}
  & P1 & 50.5  & 0.283 & 0.298 & 0.586 \\
  & P2 & 64.9 & 0.365 & 0.508 & 0.408 \\
  & P3 & 33.6 & 0.126 & 0.00 & 0.721 \\
\bottomrule
\end{tabular}
\end{table}

\begin{figure}[t]
    \centering
    \includegraphics[width=\linewidth]{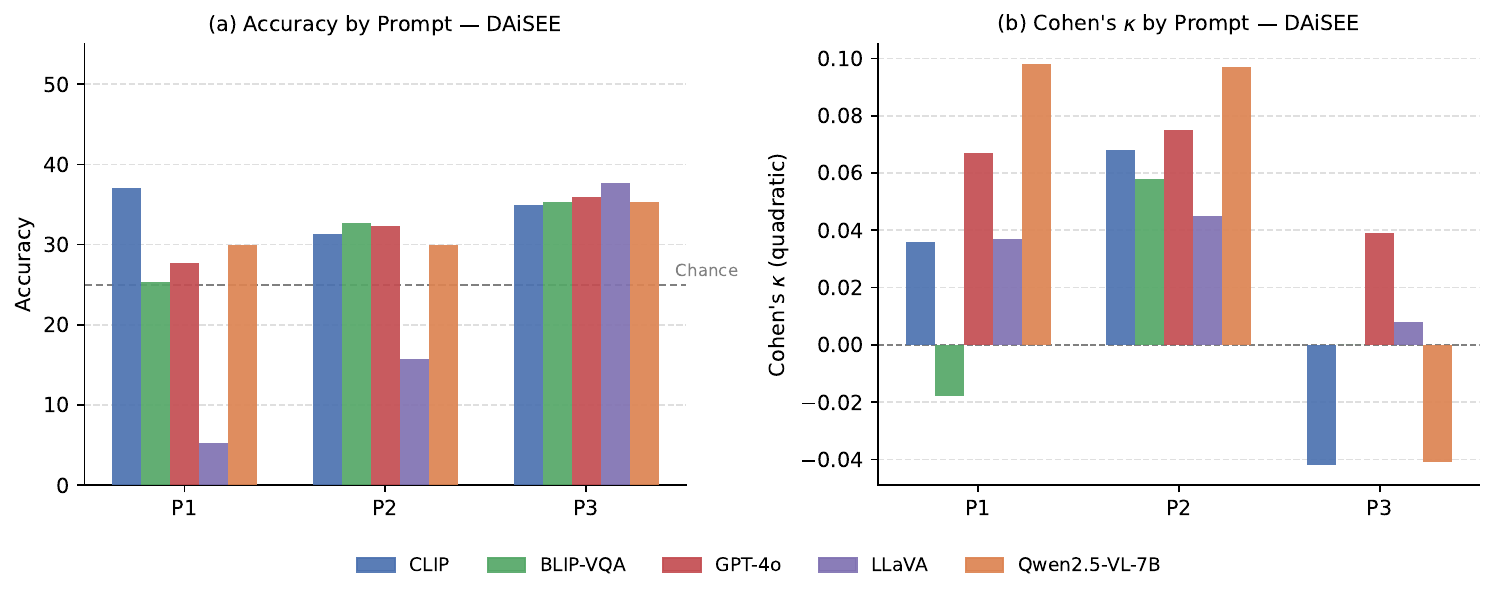}
    \caption{Accuracy and Cohen's $\kappa$ across the three prompt
    variants on DAiSEE.  LLaVA's accuracy spans from 5.3\% (P1)
    to 39.0\% (P3)---a 33.7 pp swing on identical images.  Dashed line
    in (a) marks the four-class random-chance baseline (25\%).}
    \label{fig:prompt_sensitivity}
\end{figure}

\subsection{Analysis of Failure Modes}
\label{sec:analysis}

\begin{figure}[t]
    \centering
    \includegraphics[width=\linewidth]{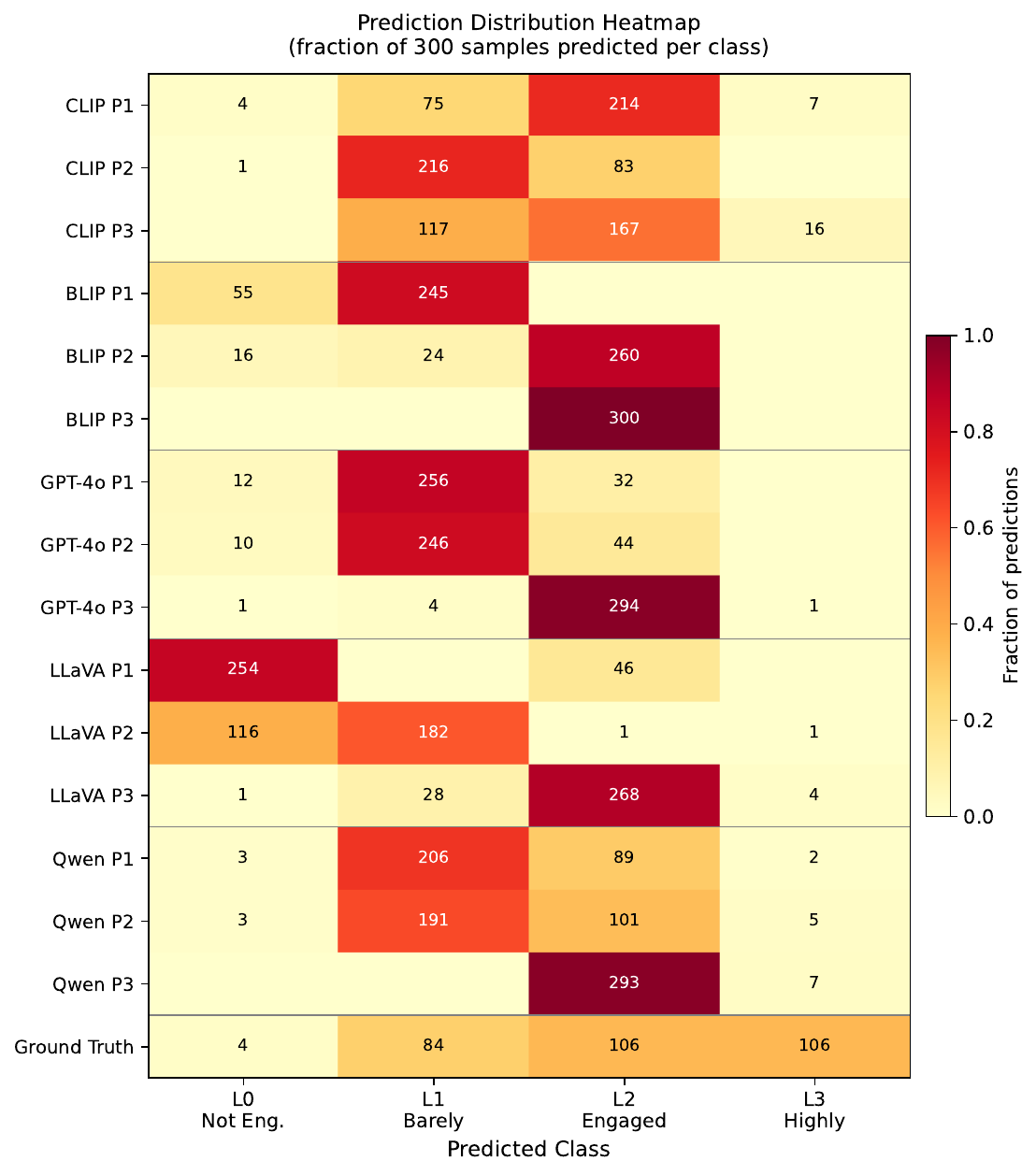}
    \caption{Fraction of the 300 DAiSEE samples predicted at each
    engagement level, for every model-prompt combination and the ground
    truth distribution.  Warm colours indicate high concentration.
    Each model collapses to 1--2 dominant classes, confirming class
    collapse rather than genuine multi-class discrimination.}
    \label{fig:heatmap}
\end{figure}

\paragraph{Class collapse.}
Figure~\ref{fig:heatmap} illustrates the predicted class distributions.
BLIP-VQA with P3 assigns \emph{all} 300 DAiSEE samples to level 2,
achieving 35.3\% accuracy solely by matching the plurality class.
GPT-4o concentrates 82--95\% of predictions on level 1 across P1 and P2,
suggesting a systematic low-engagement bias.  No model ever assigns more
than 5\% of samples to level 3 (highly engaged) despite it comprising
35\% of the ground truth.  This asymmetric collapse is not random noise;
it reflects each model's learned prior over language space that
``engaged'' or ``barely engaged'' are more plausible descriptions of a
sitting student than ``highly engaged.''

\begin{figure}[t]
    \centering
    \includegraphics[width=\linewidth]{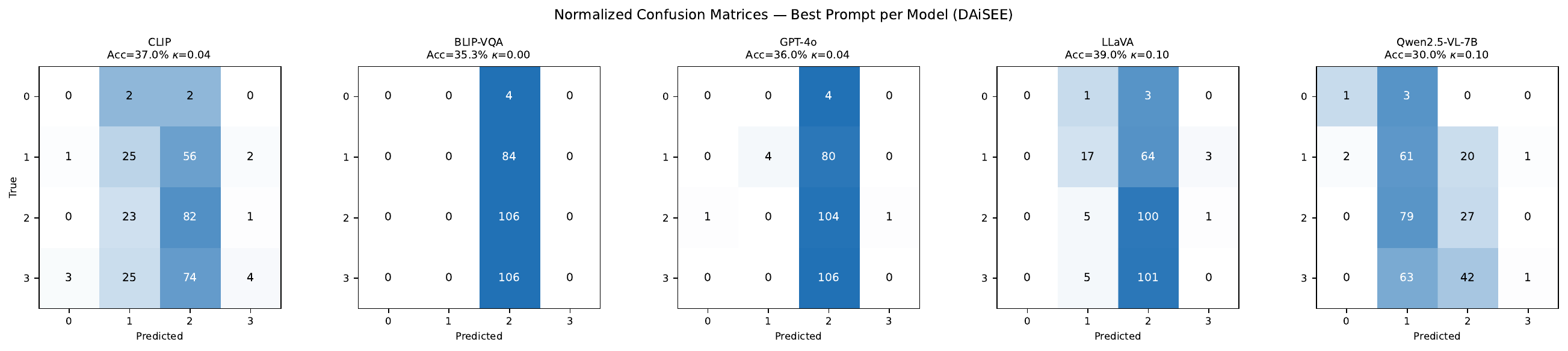}
    \caption{Normalised confusion matrices (rows = true label, cols =
    predicted label) for the best prompt per model on DAiSEE.
    Cell values show raw counts.  All models achieve near-zero recall on
    levels 0 and 3, confirming class collapse.}
    \label{fig:confusion}
\end{figure}

\paragraph{Confusion matrix analysis.}
Figure~\ref{fig:confusion} shows normalised confusion matrices for the
best configuration of each model.  All models achieve reasonable recall
on level 2 (the largest class) but near-zero recall on levels 0 and 3.
LLaVA shows the broadest distribution, occasionally predicting level 1,
but still fails on level 3 entirely.  The diagonal patterns confirm that
apparent accuracy gains are driven by level-2 majority-class capture
rather than calibrated multi-level discrimination.

\paragraph{Prompt sensitivity.}
Table~\ref{tab:sensitivity} quantifies the range of each model's
performance across the three prompts.  LLaVA exhibits the highest
sensitivity: a 32.3 pp accuracy swing on DAiSEE and 3.31 MSE variation.
On SCB the effect is even more pronounced for CLIP (53 pp swing from P1
to P3).  BLIP-VQA is the most stable model---but for the wrong reason:
it is insensitive because it \emph{always} predicts the same label
regardless of prompt or image.

\begin{table}[t]
\centering
\small
\caption{Prompt sensitivity (max $-$ min across P1/P2/P3) for each model
  and dataset.  Higher values indicate greater dependence on prompt phrasing.}
\label{tab:sensitivity}
\setlength{\tabcolsep}{4pt}
\begin{tabular}{@{}lcccccc@{}}
\toprule
& \multicolumn{3}{c}{DAiSEE} & \multicolumn{3}{c}{SCB} \\
\cmidrule(lr){2-4}\cmidrule(lr){5-7}
Model & $\Delta$Acc & $\Delta\kappa$ & $\Delta$MSE & $\Delta$Acc & $\Delta\kappa$ & $\Delta$MSE \\
\midrule
CLIP     & 5.7  & 0.110 & 0.389 & 52.9 & 0.499 & 1.978 \\
BLIP-VQA & 10.0 & 0.076 & 1.703 & 0.0  & 0.000 & 0.000 \\
GPT-4o   & 8.3  & 0.060 & 0.987 & 28.8 & 0.428 & 0.316 \\
LLaVA    & 32.3 & 0.037 & 3.313 & 41.8 & 0.251 & 4.581 \\
\bottomrule
\end{tabular}
\end{table}

\paragraph{Self-consistency.}
We ran the best prompt three times with temperature $T\!=\!0.7$ for
GPT-4o (P2) and LLaVA (P3).  CLIP and BLIP-VQA are deterministic and
trivially achieve 100\% agreement.  GPT-4o achieves full 3-run agreement
on only 68.0\% of images (pairwise: 79.0\%), while LLaVA achieves 73.3\%
(pairwise: 82.2\%).  Kappa across LLaVA runs spans from 0.008 to 0.101,
meaning a single run's reported kappa is unreliable as an estimate of
model quality.  GPT-4o additionally triggered intermittent safety refusals
in run 3 that did not occur in runs 1 or 2, further degrading consistency.

\paragraph{GPT-4o safety guardrails.}
Chain-of-thought prompting (P3) asks the model to describe ``the
student's facial expression,'' which GPT-4o interprets as facial
recognition of a real person.  This triggers its safety filters in
$\approx\!98\%$ of DAiSEE images: the model responds with ``I'm unable
to assist with requests involving identifying or describing individuals
from their photographs.''  Our parsing pipeline falls back to level 2
for all refusals, artificially inflating P3 accuracy to 36\% while
providing zero true information.  This is a practical barrier for
any system deploying GPT-4o as an individual student observer.
No such refusals occur with the scene-level SCB images.

\subsection{Feature Space Analysis}
\label{sec:feature_analysis}

To understand the failure modes mechanistically, we analyse CLIP's internal
representations directly, extracting ViT-B/32 image embeddings for all 300
DAiSEE frames and examining both feature geometry and text-image similarity
structure.

\begin{figure}[t]
    \centering
    \includegraphics[width=\linewidth]{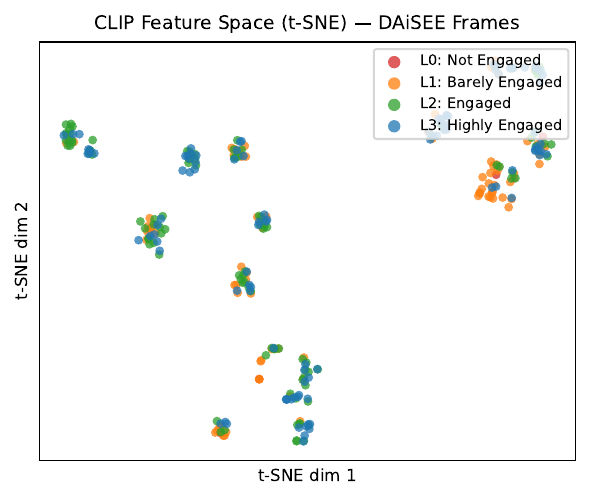}
    \caption{t-SNE projection of CLIP ViT-B/32 image embeddings for all 300
    DAiSEE frames, coloured by ground-truth engagement label.  The four
    engagement classes intermix within every cluster, confirming that CLIP
    organises frames by subject identity rather than engagement state.}
    \label{fig:tsne}
\end{figure}

\paragraph{Feature geometry.}
Figure~\ref{fig:tsne} shows a t-SNE projection of image embeddings coloured
by ground-truth engagement label.  Frames cluster by \emph{student identity
and visual appearance} rather than engagement level---all four engagement
classes intermix within every cluster.  CLIP organises the representation
space around ``who this person is'' (lighting, face, background) rather than
``how engaged they are.''  This provides a direct geometric explanation for
why zero-shot classification fails: the engagement signal does not correspond
to a separable direction in CLIP's feature space.

\begin{figure}[t]
    \centering
    \includegraphics[width=\linewidth]{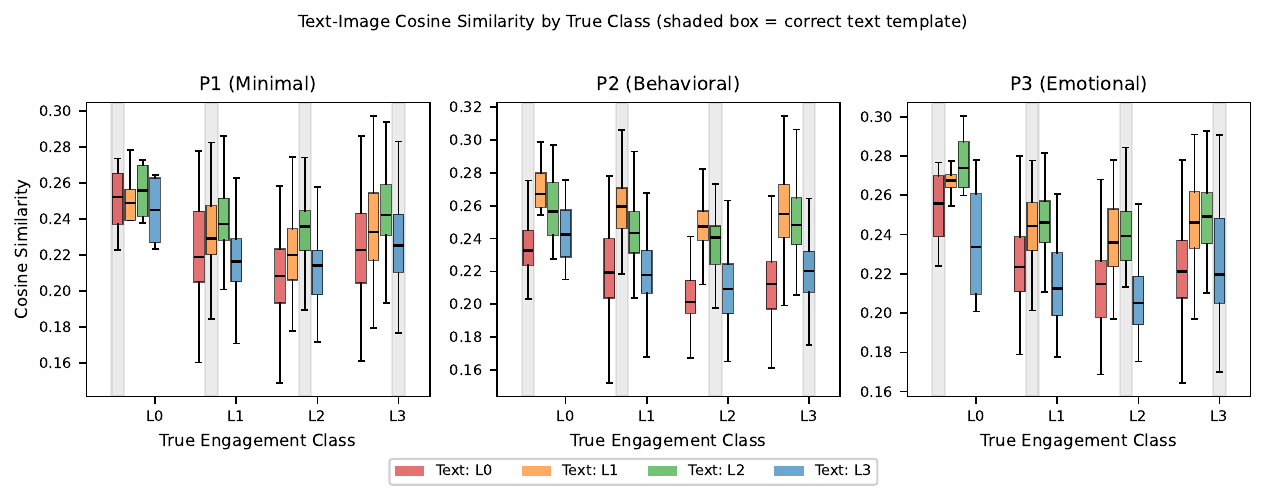}
    \caption{Text-image cosine similarity distributions across all three
    prompt variants, stratified by true engagement class.  Each group of four
    boxes shows the similarity to text templates L0--L3; the shaded box marks
    the \emph{correct} template.  Distributions overlap heavily across all
    prompts and the correct template never achieves the highest
    median---confirming that CLIP has no discriminative signal for individual
    engagement level.}
    \label{fig:similarity}
\end{figure}

\paragraph{Text-image similarity collapse.}
Figure~\ref{fig:similarity} shows cosine similarities between image embeddings
and each of the four class text templates, stratified by true engagement class,
across all three prompt variants.  Two observations are consistent across all
prompts: (1)~the correct text template (shaded column) does not yield the
highest similarity for any true class---the diagonal is never dominant;
(2)~the total similarity range spans only $\approx\!0.07$, meaning all four
text embeddings are nearly equidistant from every image.  In this regime,
argmax predictions are determined by which text template happens to land
marginally closer in embedding space, irrespective of image content.
Concretely, for P1 the L2 template wins globally, collapsing all predictions
to ``Engaged''; for P2, L1 dominates; for P3, L1 and L2 differ by
$<\!0.003$, producing high run-to-run variance.  This directly explains both
the class collapse and the extreme prompt sensitivity reported in
Tables~\ref{tab:daisee_allprompts} and~\ref{tab:sensitivity}.

%% file: sec/discussion.tex
\section{Discussion}
\label{sec:discussion}

\paragraph{Why does scene-level outperform individual-level?}
The performance gap between DAiSEE ($\kappa \leq 0.10$) and SCB
($\kappa$ up to 0.60) is not simply a function of dataset difficulty.
We hypothesise three contributing factors.  First, \emph{behavioural
salience}: SCB labels derive from overt physical actions (raised hand,
phone use) that are visually unambiguous to a language-grounded model,
whereas DAiSEE labels reflect internal cognitive states that manifest
only subtly in facial microexpressions.  Second, \emph{context richness}:
a classroom scene provides spatial layout, group dynamics, and object
context (\eg, desks, boards) that help ground engagement inference,
while a webcam close-up of a single face offers minimal contextual anchors.
Third, \emph{label construction}: the SCB aggregation scheme maps
explicit physical categories to ordinal scores, creating a more
objective signal than subjective per-clip engagement ratings.

\paragraph{Implications for system design.}
Our findings suggest that deploying VLMs as individual student observers
in their current zero-shot form is premature.  A more productive
paradigm may be \emph{scene-first analysis}: aggregate visual information
at the classroom level, then query a VLM.  This approach aligns better
with how teachers actually monitor classrooms---by scanning the room
rather than studying individual faces.  For privacy reasons, scene-level
analysis is also preferable as it avoids close facial analysis of minors.

\paragraph{Prompt engineering as unreliable compensation.}
The wide prompt sensitivity we observe means that practitioners who tune
prompts on a small validation set risk over-fitting to that distribution.
A prompt that yields 67.9\% accuracy for GPT-4o on SCB could easily
degrade by 28 pp on a different class distribution.  Reliable deployment
requires either task-specific fine-tuning or principled prompt selection
with held-out validation.

\paragraph{Open vs. closed source.}
LLaVA-1.5-7B (4-bit quantised, running locally) matches or outperforms
GPT-4o on DAiSEE, and achieves competitive performance on SCB at zero
API cost and without data leaving the institution.  This is
encouraging for privacy-sensitive educational deployments, though the
comparison is not perfectly controlled given the quantisation difference.

\paragraph{Limitations.}
This study is limited to zero-shot evaluation.  Fine-tuning VLMs on even
modest task-specific data would likely close the gap substantially.
The single-frame assumption discards temporal dynamics known to carry
engagement information.  The SCB engagement labels are derived from
behaviour categories rather than direct human ratings, which may not
fully align with teachers' engagement judgements.  Finally, results on
16 subjects in a webcam setting (DAiSEE) may not generalise to diverse
classroom environments.

\paragraph{Understanding the gaps.}
The feature space analysis (Sec.~\ref{sec:feature_analysis}) surfaces three
fundamental gaps between current VLMs and individual engagement recognition.
The \emph{representation gap}: CLIP organises visual features around
appearance-level cues---identity, lighting, background---because such cues
dominate internet-scale image--caption training data, not engagement states.
The \emph{granularity gap}: the four-class engagement scale requires
fine-grained ordinal discrimination that zero-shot VLMs are not designed
for; the $\approx\!0.07$ cosine similarity range confirms that CLIP treats
engagement levels as nearly synonymous in language space.  The
\emph{temporal gap}: engagement is inherently dynamic, yet our single-frame
protocol discards the most informative temporal signal.  Taken together,
these gaps explain why near-random performance is not an artefact of prompt
choice or dataset noise, but a structural consequence of the pre-training
objective mismatch.

\paragraph{Future directions.}
These gaps suggest concrete research paths.
\emph{Supervised adaptation}: linear probing or lightweight adapter layers
trained on even small labelled engagement datasets could reveal whether
engagement signal exists in frozen CLIP features but is entangled with
identity.
\emph{Contrastive disentanglement}: a projection that explicitly separates
engagement from identity---positive pairs sharing engagement label across
different students, negatives sharing identity across engagement
levels---would directly address the clustering-by-identity failure mode we
observe.
\emph{Temporal VLMs}: video-native models should be evaluated on full
10-second DAiSEE clips; our results motivate this as a priority, given the
insufficient signal in static frames.
\emph{Scene-to-individual transfer}: since scene-level recognition is
tractable ($\kappa \approx 0.60$), future work could explore top-down
approaches that use scene-level estimates as a prior to refine
individual-level inference.
\emph{Benchmark standardisation}: prompt sensitivity ($\Delta\kappa$) should
be reported as an explicit evaluation axis in future engagement benchmarks,
rather than a single-prompt result, to avoid over-optimistic reporting.

%% file: sec/conclusion.tex
\section{Conclusion}
\label{sec:conclusion}

We presented a systematic zero-shot benchmark of four Vision-Language
Models for classroom engagement recognition, evaluated across two
datasets and three prompt designs.  Our results establish that current
VLMs cannot reliably classify individual student engagement---with
Cohen's $\kappa$ never exceeding 0.10 on DAiSEE---but show substantially
more promise at the scene level, where CLIP and GPT-4o reach
$\kappa \approx 0.60$ on SCB.  We characterise three failure modes
(near-random ordinal agreement, class collapse, and extreme prompt
sensitivity) and document a practical deployment barrier (GPT-4o safety
refusals for individual face analysis).  We hope these baselines serve
as a rigorous starting point for the community, and that the identified
failure modes motivate targeted improvements: scene-level pre-aggregation,
prompt-robust fine-tuning, and temporal modelling to go beyond
single-frame analysis.

%% file: sec/suppl.tex
\clearpage
\setcounter{page}{1}
\maketitlesupplementary

\section{Full Prompt Texts}
\label{sec:suppl_prompts}

This supplementary provides the exact prompt text used for each model and dataset
in our zero-shot benchmark, enabling full reproducibility.

\newcommand{\promptbox}[2]{%
  \vspace{0.4em}
  \noindent\textbf{#1}\\[2pt]
  \noindent\fbox{\begin{minipage}{0.96\linewidth}
    \small\ttfamily #2
  \end{minipage}}
  \vspace{0.4em}
}

\subsection{SCB Dataset — GPT-4o, LLaVA-1.5, and Qwen2.5-VL-7B}
\label{sec:suppl_scb_prompts}

These prompts were applied to the scene-level SCB images for GPT-4o, LLaVA-1.5,
and Qwen2.5-VL-7B.
Each image was submitted directly; the model returned a rating based on visible
student behaviours.

\promptbox{P1 — Minimal}{%
Given this image, rate the overall classroom's engagement level based on the \\
majority of students:\\[4pt]
0 = not engaged\\
1 = barely engaged\\
2 = engaged\\
3 = highly engaged\\[4pt]
Answer with just the number (0, 1, 2, or 3).%
}

\promptbox{P2 — Rubric-Anchored}{%
You are an educational observer. Assess the overall classroom's engagement based \\
on the predominant student behaviors and body language:\\[4pt]
0 = Distracted and disinterested (e.g., using a phone, bowing the head)\\
1 = Passively present but not focused (e.g., leaning over the table)\\
2 = Attentive and following along (e.g., reading, writing)\\
3 = Actively focused and alert (e.g., raising a hand)\\[4pt]
Respond with only the number (0, 1, 2, or 3).%
}

\promptbox{P3 — Chain-of-Thought}{%
Analyze the students in this classroom environment to assess their collective \\
learning engagement.\\[4pt]
Step 1 (Behaviors): Describe the predominant activities of the majority of the \\
students in one detailed sentence. Focus specifically on how they are interacting \\
with their environment (e.g., writing, reading, resting heads, using devices, \\
raising hands).\\[4pt]
Step 2 (Posture \& Focus): Describe the general body posture and visual attention \\
of the class in one detailed sentence (e.g., sitting upright, leaning forward, \\
slumped over, gaze directed at desks or away).\\[4pt]
Step 3 (Rating): Based strictly on the behaviors and posture described in Steps 1 \\
and 2, rate the overall engagement level of the majority of students:\\
0 = not engaged at all\\
1 = barely engaged\\
2 = engaged\\
3 = highly engaged\\[4pt]
Format your response strictly as:\\
Behaviors: [Your sentence]\\
Posture: [Your sentence]\\
Rating: [0, 1, 2, or 3]%
}

\subsection{SCB Dataset — CLIP Text Templates}
\label{sec:suppl_scb_clip_prompts}

For CLIP (ViT-B/32) on SCB, classification uses cosine similarity between the
scene image embedding and per-class text template embeddings.

\begin{table}[h]
\centering
\small
\caption{CLIP text templates for SCB scene-level engagement classification.}
\label{tab:scb_clip_templates}
\setlength{\tabcolsep}{4pt}
\begin{tabular}{@{}llp{6.8cm}@{}}
\toprule
Prompt & Level & Text Template \\
\midrule
\multirow{4}{*}{P1} & L0 & \textit{a classroom where the majority of students are not engaged at all} \\
                    & L1 & \textit{a classroom where the majority of students are barely engaged} \\
                    & L2 & \textit{a classroom where the majority of students are engaged} \\
                    & L3 & \textit{a classroom where the majority of students are highly engaged} \\
\midrule
\multirow{4}{*}{P2} & L0 & \textit{a classroom where students are distracted, bowing heads, using phones, or completely disengaged} \\
                    & L1 & \textit{a classroom where students are passively present but not focused, leaning heavily on desks} \\
                    & L2 & \textit{a classroom where students are attentive, reading, writing, or following along} \\
                    & L3 & \textit{a classroom where students are actively focused, raising hands, showing curiosity} \\
\midrule
\multirow{4}{*}{P3} & L0 & \textit{a classroom with bored, disinterested students showing disengaged posture} \\
                    & L1 & \textit{a classroom with indifferent students showing neutral, passive posture} \\
                    & L2 & \textit{a classroom with attentive, interested students showing focused posture} \\
                    & L3 & \textit{a classroom with excited, curious students showing highly alert posture} \\
\bottomrule
\end{tabular}
\end{table}

\subsection{DAiSEE Dataset — GPT-4o, LLaVA-1.5, and Qwen2.5-VL-7B}
\label{sec:suppl_daisee_generative_prompts}

Analogous prompts were used for the individual-student DAiSEE frames.
The wording was adapted from scene-level to individual-level assessment.

\promptbox{P1 — Minimal}{%
Given this image of a student, rate their engagement level:\\[4pt]
0 = not engaged\\
1 = barely engaged\\
2 = engaged\\
3 = highly engaged\\[4pt]
Answer with just the number (0, 1, 2, or 3).%
}

\promptbox{P2 — Rubric-Anchored}{%
You are an educational observer. Assess this student's engagement based on their \\
visible behavior and body language:\\[4pt]
0 = Distracted and disinterested (e.g., looking away from the screen, yawning)\\
1 = Passively present but not focused (e.g., occasionally glancing at the screen)\\
2 = Attentive and following along (e.g., watching the screen, maintaining eye contact)\\
3 = Actively focused and alert (e.g., leaning forward, showing curiosity)\\[4pt]
Respond with only the number (0, 1, 2, or 3).%
}

\promptbox{P3 — Chain-of-Thought}{%
Analyze the student in this image to assess their learning engagement.\\[4pt]
Step 1 (Behaviors): Describe what the student appears to be doing in one sentence.\\
Step 2 (Posture \& Focus): Describe the student's body posture and where their \\
gaze appears directed in one sentence.\\
Step 3 (Rating): Based on the above, rate the student's engagement level:\\
0 = not engaged at all\\
1 = barely engaged\\
2 = engaged\\
3 = highly engaged\\[4pt]
Format your response strictly as:\\
Behaviors: [Your sentence]\\
Posture: [Your sentence]\\
Rating: [0, 1, 2, or 3]%
}

\subsection{DAiSEE Dataset — CLIP Text Templates}
\label{sec:suppl_daisee_clip_prompts}

For CLIP (ViT-B/32) on DAiSEE, classification uses cosine similarity between the
individual frame embedding and per-class text template embeddings. The class whose
template yields the highest similarity is the prediction.

\begin{table}[h]
\centering
\small
\caption{CLIP text templates for DAiSEE engagement classification.}
\label{tab:clip_templates}
\setlength{\tabcolsep}{4pt}
\begin{tabular}{@{}llp{6.8cm}@{}}
\toprule
Prompt & Level & Text Template \\
\midrule
\multirow{4}{*}{P1} & L0 & \textit{a photo of a student who is not engaged at all, completely distracted} \\
                    & L1 & \textit{a photo of a student who is barely engaged, passively present} \\
                    & L2 & \textit{a photo of a student who is engaged and attentive} \\
                    & L3 & \textit{a photo of a student who is highly engaged, very focused and alert} \\
\midrule
\multirow{4}{*}{P2} & L0 & \textit{a student looking away from the screen, yawning or distracted, showing no interest} \\
                    & L1 & \textit{a student passively sitting, occasionally glancing at the screen, low attention} \\
                    & L2 & \textit{a student watching the screen attentively, maintaining eye contact with the content} \\
                    & L3 & \textit{a student leaning forward, deeply focused, actively concentrating on the content} \\
\midrule
\multirow{4}{*}{P3} & L0 & \textit{a bored and disinterested student with a blank or sleepy expression} \\
                    & L1 & \textit{a slightly disengaged student with a neutral, unfocused expression} \\
                    & L2 & \textit{an attentive student with an interested and alert facial expression} \\
                    & L3 & \textit{a highly focused student showing curiosity and active concentration} \\
\bottomrule
\end{tabular}
\end{table}

\subsection{BLIP-VQA Question Strings}
\label{sec:suppl_blip_prompts}

BLIP-VQA receives a question string alongside the image. The free-text answer
is mapped to an engagement level via a hand-crafted lookup table.

\promptbox{P1 — Minimal}{%
How engaged is this student? Answer: not engaged, barely engaged, engaged, or\\
highly engaged.%
}

\promptbox{P2 — Rubric-Anchored}{%
Based on the student's visible behavior, are they: (0) distracted and disinterested,\\
(1) passively present, (2) attentive and following along, or (3) actively focused?\\
Answer with 0, 1, 2, or 3.%
}

\promptbox{P3 — Chain-of-Thought}{%
Describe the student's facial expression and body posture, then rate their engagement\\
as: not engaged, barely engaged, engaged, or highly engaged.%
}